\pdfoutput=1

\documentclass[11pt]{article}

\usepackage{acl}

\usepackage{times}
\usepackage{latexsym}

\usepackage[T1]{fontenc}

\usepackage[utf8]{inputenc}

\usepackage{microtype}

\usepackage{graphicx}
\usepackage{amsmath}
\usepackage{pgfplots}
\usepackage{subfig}
\usepackage{multirow}
\usepackage{soul}

\title{Who Are We Talking About? \\ Handling Person Names in Speech Translation}

\author{Marco Gaido$^{1,2}$,  \textbf{Matteo Negri}$^1$ and \textbf{Marco Turchi}$^1$ \\
  $^1$Fondazione Bruno Kessler \\
  $^2$University of Trento \\
  Trento, Italy \\
  {\tt \{mgaido,negri,turchi\}@fbk.eu} \\}

\begin{document}
\maketitle
\begin{abstract}

Recent work has shown that systems for speech translation (ST) -- similarly to automatic speech recognition (ASR) -- poorly handle person names. This shortcoming does not only lead to errors that can seriously distort the meaning of the input, but also hinders the adoption of such systems in application scenarios (like computer-assisted interpreting) where the translation of named entities, like person names, is crucial. In this paper, we first analyse the outputs of ASR/ST systems to identify the reasons of failures in person name transcription/translation. Besides the frequency in the training data, we pinpoint the nationality of the referred person as a key factor. We then mitigate the problem by creating multilingual models, and further improve our ST systems by forcing them to jointly generate transcripts and translations, prioritising the former over the latter. Overall, our solutions result in a relative improvement in token-level person name accuracy by 47.8\% on average for three language pairs (en$\rightarrow$es,fr,it).

\end{abstract}

\section{Introduction}

Automatic speech translation (ST) is the task of generating
the textual translation of utterances.
Research on ST \cite{anastasopoulos-etal-2021-findings,bentivogli-etal-2021-cascade} has so far focused on 
comparing the \textit{cascade} (a pipeline of an automatic speech recognition -- ASR -- and a machine translation -- MT -- model) and
\textit{direct} paradigms \citep{berard_2016,weiss2017sequence}, or on improving either of them in terms of overall quality.
Quality is usually measured with automatic metrics such as BLEU \citep{papineni-2002-bleu} and TER \citep{snover-ter-2006}, possibly corroborated by manual analyses.

These metrics -- as well as neural-based ones like COMET \citep{rei-etal-2020-comet} -- are relatively insensitive
to errors on named entities (NEs) and numbers \citep{Amrhein2022IdentifyingWI},
which instead are of paramount importance
for human readers \citep{Xie2022}.
As such, the blind pursue of higher scores can lead to systems biased toward the metrics and not targeted
on real users.

In addition, there are cases in which users are interested only in NEs.
For instance, interpreters easily craft more fluent and intelligible translations than machines \citep{fantinuoli-prandi-2021-towards}, but during simultaneous sessions suffer from a high cognitive workload  \citep{bianca_prandi_2018_1493293,Desmet_2018_simultaneous_interpretation}, to which
NEs and specific terminology significantly contribute  \citep{jones-1998-interpreting,gile-2009-basic,bianca_prandi_2018_1493293,Desmet_2018_simultaneous_interpretation}.
Indeed, these elements \textit{i)} are hard to remember \cite{liu_2004_working_memory}, \textit{ii)} can be unknown to interpreters and difficult to recognize \cite{GRIFFIN1998313}, and \textit{iii)} differently from other types of words, usually have one or few correct translations.
For this reason, modern computer-assisted interpreting (CAI -- \citealt{fantinuoli-2017-cai}) tools aim at automatically recognizing, displaying, and \textit{translating} NEs and terms. However, current solutions rely on pre-defined dictionaries to identify and translate the elements of interest \citep{Fantinuoli2022KUDOIA}, preventing them to both generalize and disambiguate homophones/homonyms. This would be instead possible using ST system, but they need to reliably recognize and translate NEs and terms, without generating wrong suggestions that are even harmful \cite{stewart-etal-2018-automatic}.

In contrast with these needs,
\citet{gaido2021moby} recently showed on their newly created benchmark -- NEuRoparl-ST --
that both ASR models (and thus cascade ST systems) and direct ST systems
perform poorly on person names, with transcription/translation accuracy of \textasciitilde40\%.
Hence, as a first step toward ST systems more targeted for human needs, and in particular toward
the long-term goal of integrating ST models in assistant tools for live interpreting, this work focuses on \textit{i)} identifying the factors that lead to the wrong transcription and translation of person names, and \textit{ii)} proposing dedicated solutions to mitigate the problem.

To achieve these  objectives, our first contribution ($\S$\ref{subsec:expsettings}) 
is the annotation\footnote{Available at: \url{https://ict.fbk.eu/neuroparl-st/}.}
of each person name occurring in NEuRoparl-ST with information about 
their
nationality and the nationality of the speaker (as a proxy of 
the native language)
--
e.g. if a German person
says
``\textit{\underline{Macron} is the French president}'', the speaker nationality is German, while the referent nationality is 
French.
Drawing on this additional information, our second contribution
($\S$\ref{subsec:rolefreq}-\ref{subsec:person_country})
is the analysis of the concurring factors involved in the correct recognition of  person names.
Besides their frequency, we identify
as key discriminating factor  the presence in the training data of speech uttered in the referent's native language (e.g. French in the above example).
This finding, together with an observed accuracy gap between person name transcription (ASR) and translation (ST), leads to  our third contribution ($\S$\ref{sec:impro}): 
a multilingual ST system that jointly transcribes and translates the input audio,
giving higher importance to the transcription task 
in favour of a more accurate translation of names.
Our
model shows relative 
gains
in person name translation by 48\% on average 
on
three language pairs (en$\rightarrow$es,fr,it),
producing useful translations for interpreters in 66\% of the cases.

\section{Related Work}


When the source modality is text, person names can often be ``copied'', i.e. replicated unchanged, into the output. 
This task has been shown to be well accomplished by both statistical and neural translation systems \cite{koehn-knowles-2017-six}.
On the contrary, 
when the source modality is speech (as in ASR and ST), systems struggle 
due to the impossibility to copy the audio source.
The recognition of person names from speech is a complex task that has mostly been studied in the context of recognizing a name from a pre-defined list, such as phone contacts \citep{raghavan-allan-2005-matching,Suchato_2011,Bruguier_2016}.
The scenario of an open or undefined set of possible names is instead under-explored.
Few studies \citep{ghannay2018endtoend,caubriere-etal-2020-named} 
focus on comparing end-to-end and cascade approaches in the transcription and recognition of NEs from speech.
They do not directly investigate person names though, as they
do not disaggregate their results by NE category.
Similarly, \citet{Porjazovski_2021_ner_speech} explore NE recognition from speech in low-resource languages,
and propose two end-to-end 
methods: one adds a tag after each word in the generated text to define whether it is a NE or not, and one uses a dedicated decoder. 
However, they
do not provide specific insights on the system ability to 
correctly generate person names and limit their study to ASR, without investigating ST. 
Closer to our work, \citet{gaido2021moby} highlight the difficulty of 
ASR/ST
neural models to 
transcribe/translate
NEs and terminology. Although they identify person names as the hardest NE category by far, they neither analyse the root causes nor propose mitigating solutions.

\section{
Factors Influencing Name Recognition}
\label{sec:factors}

As shown in \cite{gaido2021moby}, 
the translation of person names is difficult
both for direct and cascade ST systems, which achieve similar accuracy scores (\textasciitilde40\%).
The low performance of cascade solutions is largely due to errors made by the ASR component, while the MT model usually achieves nearly perfect scores.
For this reason, henceforth we will focus on identifying the main issues related to the transcription and translation of person names, respectively in ASR and 
\textit{direct}
ST.

We hypothesize
that three main factors influence the
ability of a system to transcribe/translate a person name:
\textit{i)} its frequency in the training data, as neural models are known to poorly handle rare words, \textit{ii)} the nationality of the referent, as different languages may involve different phoneme-to-grapheme mappings and may contain different sounds, and \textit{iii)} the nationality of the speaker, as 
non-native speakers typically have different accents and hence different pronunciations of the same name.
To validate these hypotheses, we inspect the outputs of Transformer-based \citep{transformer} ASR and ST models trained with the  configuration defined in \cite{wang-etal-2020-fairseq}. 
For the sake of reproducibility, complete details on our experimental settings are provided in the Appendix.\footnote{Code available at: \url{https://github.com/hlt-mt/FBK-fairseq}.}

\subsection{Data and Annotation}
\label{subsec:expsettings}

To enable fine-grained evaluations on the three factors we suppose to be influential, we enrich the NEuRoparl-ST benchmark 
by adding three (one for each factor) features to each token annotated as \textit{PERSON}.
These are:
\textit{i)} 
the token
frequency in the target transcripts/translations of the training set,
\textit{ii)} the nationality of the referent,
and \textit{iii)} the nationality of the speaker.
The nationality of the referents
was manually collected  by the authors through  online searches.
The nationality of the speakers, instead, was 
automatically extracted from the personal data listed in LinkedEP \cite{linkedep} using the country they represent in the European Parliament.\footnote{\label{foot:ep_speakers}
For each speech in Europarl-ST,  the speaker is referenced by link to
LinkedEP.}
All our 
systems
are trained  on Europarl-ST \cite{jairsan2020a} and MuST-C \cite{MuST-Cjournal}, 
and
evaluated on 
this
new extended version of  NEuRoparl-ST.

\subsection{The Role of Frequency}
\label{subsec:rolefreq}

As a first 
step in our analysis,
we automatically 
check how the three features added to each 
\textit{PERSON}
token correlate with the correct generation of the token itself.
Our aim is to
understand the importance of these factors and to identify interpretable reasons behind the correct or wrong handling of person names.
To this end, we train a classification decision tree \cite{breiman1984classification}. Classification trees recursively divide the dataset into two groups, choosing a feature and a threshold that minimize the entropy of the resulting groups with respect to the target label. As such, they do not assume a linear relationship between the input and the target 
(like multiple regression and random linear mixed effects do)
and are a good fit for categorical features as most of ours are.
Their structure makes them easy to interpret \cite{Wu2008}:  the first decision (the root of the tree) is the most important criterion according to the learned model, while less discriminative features are pushed to the bottom.

We feed the classifier with 49 features, corresponding to: \textit{i)} the frequency of the token in the training data, \textit{ii)} the one-hot encoding of the speaker nationality, and \textit{iii)} the one-hot encoding of the referent nationality.\footnote{Speakers and referents respectively belong to 17 and 31 different nations.}
We then train it
to predict whether our ASR model is able to correctly transcribe the token in the output. 
To this end, we use the implementation of scikit-learn \citep{scikit-learn}, 
setting to 3 the maximum depth of the tree, and using Gini index as entropy measure.

Unsurprisingly, the root node decision is based on the frequency of the token in the training data, with 2.5 as  split value.
This means that person names occurring at least 3 times in the training data are likely to be correctly handled by the models.
Although this threshold may vary across datasets of different size, it is an indication on the necessary number of occurrences of a person name, eventually useful for data augmentation techniques aimed at exposing the system to relevant instances at training time (e.g. names of famous people in the specific domain of a talk to be translated/interpreted).
We validate that this finding also holds for ST systems by reporting in Table \ref{tab:3freq}
the accuracy of person tokens for ASR and the three ST language directions, 
split according to the mentioned threshold of frequency in the training set.
On average, names occurring at least 3 times in the training set are correctly generated in slightly more than 50\% of the cases, 
a much larger value compared to those with less than 3 occurrences.

\begin{table}[h]
\centering
\small
\begin{tabular}{l|ccc}
 & \textbf{All} & \textbf{Freq. >= 3} & \textbf{Freq. < 3} \\
 \hline
 \textbf{ASR} & 38.46 & 55.81 & 4.55 \\
 \textbf{en-fr} & 28.69 & 45.45 & 0.00 \\
 \textbf{en-es} & 35.29 & 53.57 & 19.05 \\
 \textbf{en-it} & 29.70 & 46.77 & 2.56 \\
 \hline
 \textbf{Average} & 33.04 & 50.40 & 6.54 \\
\end{tabular}
\caption{Token-level accuracy of person names divided into two groups according to their frequency in the training set for ASR and ST 
(en$\rightarrow$es/fr/it) systems.}
\label{tab:3freq}
\end{table}

The other nodes of the classification tree contain less interpretable criteria, which can be considered as spurious cues.
For instance, at the second level of the tree,
a splitting criterion is ``\textit{is the speaker from 
Denmark?}'' because the only talk by a Danish speaker contains a mention to \textit{Kolarska-Bobinska} that systems were not able to correctly generate. 

We hence decided to perform further dedicated experiments to better understand the role of the the other two factors: referent and speaker nationality.

\subsection{The Role of Referent Nationality}
\label{subsec:person_country}

Humans often struggle to understand names belonging to languages that are different from their 
native one or from those
they know. Moreover,
upon manual inspection of the system outputs, we observed that some names 
were
Englishized (e.g. \textit{Youngsen} instead of \textit{Jensen}). In light of this, we 
posit
that a system trained to recognize English sounds and to learn
English phoneme-to-grapheme mappings might be inadequate to handle non-English names.

We first validate this idea by computing the accuracy for names of people from the United Kingdom\footnote{
We are aware  that our annotation is potentially subject to noise, due to the
possible
presence of UK citizens with 
non-anglophone
names. 
A thorough study on the best strategies to maximise the accuracy of UK/non-UK label assignment is a task \textit{per se}, out of the scope of this work. By now, as a manual inspection of the names revealed no such cases in our data, we believe that the few possible wrong assignments do not undermine our experiments, nor the reported findings.} (`UK'' henceforth) and for names of people from the rest of the World (``non-UK'').
Looking at Table \ref{tab:subjectacc}, we notice that our assumption seems to hold for both ASR and ST. However,
the scores
correlate with the frequency (Freq.) of names in the training set\footnote{\label{foot:freq}Notice that the ASR and the ST training sets mostly contain the same data, so frequencies are similar in the four cases.} as, on average, UK referents have more than twice the occurrences (46.21) 
of
non-UK referents (21.96). 
The higher scores for UK referents may hence 
depend on
this second factor.

\begin{table}[t!]
\centering
\small
\begin{tabular}{l|ccccc}
 \textbf{Referent} & \textbf{ASR} & \textbf{en-fr} & \textbf{en-es} & \textbf{en-it} & \textbf{Freq.} \\
 \hline
 \textbf{UK} & 52.38 & 59.09 & 63.16 & 41.18 & 46.21 \\
 \textbf{non-UK} & 35.78 & 22.00 & 30.00 & 27.38 & 21.96 \\
 \hline
 \textbf{All} & 38.46 & 28.69 & 35.29 & 29.70 & 25.65 \\
\end{tabular}
\caption{Token-level accuracy of ASR and ST 
(en-fr, en-es, en-it)
systems for 
UK/non-UK \textit{referents}.}
\label{tab:subjectacc}
\end{table}

To disentangle the two factors and isolate the impact of 
referents'
nationality, we create a training set with balanced average frequency for UK and non-UK people by filtering out a subset of the instances containing UK names from the original training set.\textsuperscript{\ref{foot:ep_speakers}} 
To ensure that our results are not due to a particular filtering method, we randomly choose the instances to remove and run the experiments on three different filtered training sets.
The results for the three ST language pairs and ASR (see Table \ref{tab:samefreq_subject}) confirm the presence of a 
large
accuracy gap between UK and non-UK names (9.22 on average), meaning that the accuracy on non-UK names (23.62) is on average \textasciitilde 30\% lower than the accuracy on UK names (32.84).
As in this case we can rule out any bias 
in
the results due to the frequency in the training set, we can state that the 
nationality of the
referent is an important factor.

\begin{table}[ht]
\centering
\small
\begin{tabular}{l|ccccc}
 & \textbf{ASR} & \textbf{en-fr} & \textbf{en-es} & \textbf{en-it} & \textbf{Avg.} \\
 \hline
 \textbf{UK} & 42.86 & 25.76 & 33.33 & 29.41 & 32.84 \\
 \textbf{non-UK} & 29.05 & 22.67 & 23.33 & 19.44 & 23.62 \\
 \hline
 \textbf{$\Delta$Accuracy} & 13.81 & 3.09 & 10.00 & 9.97 & 9.22 \\
\end{tabular}
\caption{Token-level accuracy of UK/non-UK 
\textit{referents} averaged over three runs with balanced training sets.}

\label{tab:samefreq_subject}
\end{table}

\subsection{The Role of Speaker Nationality}
\label{subsec:rolesnat}

Another factor 
likely to
influence the correct understanding of person names from speech is the speaker accent. To verify its impact, we follow a similar procedure to that of the previous section. First, we check whether the overall accuracy is higher for names uttered by UK speakers than for those uttered by non-UK speakers. Then, to ascertain whether the results depend on the proportion of UK/non-UK speakers, we randomly create three training sets featuring a balanced average frequency of speakers from the two groups.

\begin{table}[ht]
\centering
\small
\begin{tabular}{l|ccccc}
 \textbf{Speaker} & \textbf{ASR} & \textbf{en-fr} & \textbf{en-es} & \textbf{en-it} & \textbf{Freq.} \\
 \hline
 \textbf{UK} & 41.03 & 32.43 & 36.84 & 29.41 & 34.55 \\
 \textbf{non-UK} & 37.36 & 27.06 & 34.57 & 29.85 & 21.76 \\
 \hline
 \textbf{All} & 38.46 & 28.69 & 35.29 & 29.70 & 25.65 \\
\end{tabular}
\caption{Token-level accuracy of ASR and ST
systems for names uttered by 
UK/non-UK
\textit{speakers}.}
\label{tab:speakeracc}
\end{table}

Table \ref{tab:speakeracc} shows the overall results split according to
the two groups of speaker nationalities. In this case, the accuracy gap is 
minimal (the maximum gap is 5.37 for en-fr, while it is even negative for en-it), suggesting that the speaker accent has marginal influence, if any,  on how ASR and ST systems handle person names.

The experiments on 
balanced training sets
(see Table \ref{tab:samefreq_speaker}) confirm the above results, with an average accuracy difference of  2.78  for ASR and the three ST language directions.
In light of this, we can conclude that, differently from the other two factors, speakers' nationality has negligible 
effects on
ASR/ST performance on person names.

\begin{table}[ht]
\centering
\small
\begin{tabular}{l|ccccc}
 \textbf{Speaker} & \textbf{ASR} & \textbf{en-fr} & \textbf{en-es} & \textbf{en-it} & \textbf{Avg.} \\
 \hline
 \textbf{UK} & 29.91 & 29.73 & 28.95 & 23.53 & 28.03 \\
 \textbf{non-UK} & 33.33 & 22.75 & 25.51 & 19.40 & 25.25 \\
 \hline
 \textbf{$\Delta$Accuracy} & -3.42 & 6.98 & 3.43 & 4.13 & 2.78 \\
\end{tabular}
\caption{Token-level accuracy of person names uttered by 
UK/non-UK
\textit{speakers} averaged over
three runs with balanced training sets.}
\label{tab:samefreq_speaker}
\end{table}

\begin{table*}[ht]
\centering
\small
\begin{tabular}{l||c|ccc||c|ccc||c}
 & \multicolumn{4}{c||}{\textbf{Monolingual}} & \multicolumn{4}{c||}{\textbf{Multilingual}} & \\ 
 & \textbf{ASR} & \textbf{en-fr} & \textbf{en-es} & \textbf{en-it} & \textbf{ASR} & \textbf{en-fr} & \textbf{en-es} & \textbf{en-it} &  \\
 \hline
 & \multicolumn{1}{c|}{\textbf{WER ($\downarrow$)}}   & \multicolumn{3}{c||}{\textbf{BLEU ($\uparrow$)}}                  & \multicolumn{1}{c|}{\textbf{WER ($\downarrow$)}}   & \multicolumn{3}{c||}{\textbf{BLEU ($\uparrow$)}} &         \\ 
 \hline
 \textbf{Europarl-ST} & 13.65 & 32.42 & 34.11 & 25.72 & 13.29 & 33.92 & 35.59 & 26.55 & \\
 \textbf{MuST-C} & 11.17 & 32.81 & 27.18 & 22.81 & 11.86 & 33.34 & 27.72 & 23.02 & \\
 \hline
    & \multicolumn{8}{c||}{\textbf{Token-level Person Name Accuracy ($\uparrow$)}}     & \textbf{Avg. $\Delta$}        \\ 
\hline
 \textbf{Overall} & 38.46 & 28.69 & 35.29 & 29.70 & 46.15 & 38.52 & 44.54 & 36.63 & +8.43 \\
 \hline
 \textbf{UK} & 52.38 & 59.09 & 63.16 & 41.18 & 66.67 & 59.09 & 63.16 & 52.94  & +6.51\\
 \textbf{non-UK} & 35.78 & 22.00 & 30.00 & 27.38 & 42.20 & 34.00 & 41.00 & 33.33 & +8.84 \\
\end{tabular}
\caption{Transcription/translation quality measured respectively with WER and SacreBLEU\footnotemark~\cite{post-2018-call} and token-level person name accuracy,
both overall and divided into 
UK/non-UK
%
referents. \textit{Avg. $\Delta$} indicates the difference between  multilingual and monolingual systems averaged over the ASR and the 
three ST directions.
}

\label{tab:multiling}
\end{table*}

\section{Improving Person Name Translation}
\label{sec:impro}
The previous section has uncovered that only two of the three
considered factors actually have a 
tangible
impact: the frequency in the training set, and the referent nationality.
The first issue can be 
tackled either by
collecting more data, or by
generating synthetic instances  \citep{alves:hal-02907053,zheng-2021-synthetic}.
Fine-tuning
the model on additional material is usually a viable
solution in the use case of assisting interpreters since, during their preparation phase, they
have access to
various sources of information
\cite{diaz-2015-preparation-interpreting}, including
recordings of previous related sessions.
Focusing on the second issue, we hereby explore \textit{i)}  the creation of 
models that are more robust to
a wider range of phonetic features and hence
to names
of different nationalities ($\S$\ref{sec:multiling}), and \textit{ii)} the design of 
solutions to close the gap between ASR and ST systems
attested by
previous work \citep{gaido2021moby} and
confirmed by
our preliminary 
results
shown in Table \ref{tab:3freq}
($\S$\ref{sec:close_gap}).

\footnotetext{\fontsize{8.9}{9}{\texttt{BLEU+c.mixed+\#.1+s.exp+tok.13a+v.1.5.0}}}

\subsection{Increasing Robustness to non-UK Referents}
\label{sec:multiling}

As illustrated in $\S$\ref{subsec:person_country}, one cause of failure of our ASR/ST models trained on English audio is the tendency to force every sound to an English-like word, distorting person names from other languages.
Consequently, we posit that a multilingual system, trained to recognize and translate speech in different languages, might be more robust and, in turn, achieve better performance on non-English names.

We test this hypothesis by training multilingual ASR and ST models that are fed with audio in different languages, and respectively produce 
transcripts and translations (into French, Italian, or Spanish in our case).
The ST training data (*$\rightarrow$es/fr/it) consists of the en$\rightarrow$es/fr/it sections of MuST-C and the \{nl, de, en, es, fr, it, pl, pt, ro\}$\rightarrow$es/fr/it sections of Europarl-ST.
Notice that, in this scenario, the English source audio constitutes more than 80\% of the total training data, as MuST-C is considerably bigger than Europarl-ST and the English speeches in Europarl-ST are about 4 times those in the other 
languages.\footnote{For instance,  in *-fr the training set amounts to 671 hours of audio, 573 (i.e. 83\%) having English
audio.}
For ASR, we use the audio-transcript pairs of the *-it training set 
defined above. 
Complete details on our experimental settings are provided in the Appendix.\textsuperscript{\ref{foot:code_release}}

We analyze the effect of including additional languages
both in terms of general quality (measured as WER for ASR, and BLEU for ST) and in terms of person name transcription/translation accuracy.
Looking at the first two rows of Table \ref{tab:multiling}, we notice that the improvements in terms of generic translation quality (BLEU) are higher on the Europarl-ST than on the MuST-C 
test set
-- most likely because the additional data belongs to the Europarl domain -- while in terms of speech recognition (WER) there is a small improvement for Europarl-ST and a small loss for MuST-C. 
Turning to person names (third line of the table),
the gains of the multilingual models (+8.43 accuracy on average) are higher and consistent between ASR and the ST language pairs.

By 
dividing the
person names into the two categories discussed in $\S\ref{subsec:person_country}$ -- 
UK and non-UK
referents -- 
the results become less consistent across language pairs.
On ST into French and Spanish, the accuracy of UK names remains constant, while there are significant gains (respectively +12 and +11) for non-UK names.
These 
improvements seem to 
support
the intuition that
models trained on more languages learn a wider range phoneme-to-grapheme mappings and so
are able to better handle non-English names.
However, the results for ASR and 
for ST into Italian seemingly contradict our hypothesis,
as they show higher improvements for UK names (\textasciitilde11-14) than for non-UK names (\textasciitilde6-7).

\begin{table*}[ht]
\centering
\small
\begin{tabular}{l|c|ccc||cccccc}
\multirow{2}{*}{\textbf{Model}} & \multirow{2}{*}{\begin{tabular}[c]{@{}c@{}}\textbf{WER} ($\downarrow$)\\\textbf{ASR}\end{tabular}} & \multicolumn{3}{c||}{\textbf{BLEU ($\uparrow$)}} & \multicolumn{6}{c}{\textbf{Person Accuracy}} \\
                       &                                                                    & \textbf{en-es}  & \textbf{en-fr}  & \textbf{en-it}  & \textbf{ASR}   & \textbf{en-es}   & \textbf{en-fr}   & \textbf{en-it}  & \textbf{ST Avg.} & \textbf{ASR-ST} \\
\hline
Base                & 13.29 & 35.86 & 33.99 & 26.80 & 46.15 & 44.54 & 38.52 & 36.63 & 39.90 & 6.25\\ 
Triangle            & 14.25 & 37.42 & 35.44 & 28.20 & 42.31 & 43.70 & 41.80 & 41.58 & 42.36 & -0.05 \\
\hspace{0.2cm} $\lambda_{ASR}$=0.8, $\lambda_{ST}$=0.2   & 13.75 & 36.48 & 34.85 & 27.30 & 47.69 & 44.54 & 43.44 & 50.50 & 46.16 & 1.53
\end{tabular}
\caption{WER (for ASR), SacreBLEU (for ST), and token-level person name accuracy computed on the NEuRoparl-ST test set.
For triangle models, 
ASR scores are computed on the transcript output of the *-it model, as throughout the paper we evaluate ASR on the English transcript of the en-it section. 
\textit{ST Avg.} is the
the average accuracy 
on the 3 language pairs 
(en$\rightarrow$es,fr,it)
and
\textit{ASR-ST} is
the difference between 
the ASR 
and the average ST accuracy.}
\label{tab:triangle}
\end{table*}

We investigate this behavior by further dividing the non-UK group into two sub-categories:
the names of referents whose native language is included in the training set
(``in-train'' henceforth),
and 
those
of referents whose native language is not included in the training set
(``out-of-train'').
For in-train non-UK names, the monolingual ASR accuracy is 33.33 and
is outperformed by the multilingual counterpart by 16.66, i.e. by a margin higher than that for UK names (14.29). For the out-of-train names, instead, the gap between the monolingual ASR accuracy (36.71) and the multilingual ASR accuracy (39.24) is marginal (2.5). 
Similarly, for ST into Italian the in-train group accuracy improves by 8.70 (from 34.78 to 43.48), while the out-of-train group accuracy has a smaller gain of 4.92 (from 24.59 to 29.51).
These results 
indicate
that adding a language to the training data helps the correct handling of person names belonging to that language, even when translating/transcribing from another language.
Further evidence is exposed in $\S\ref{sec:err_analysis}$, where we analyse the errors
made by our systems and how their distribution changes between a monolingual and a multilingual one.

\subsection{Closing the Gap Between ASR and ST}
\label{sec:close_gap}

The previous results -- in line with those of \citet{gaido2021moby} -- 
reveal
a gap between ASR and ST systems, although their task is similar when it comes to person names. Indeed, 
both ASR and ST 
have
to recognize 
the
names
from the speech, and produce 
them
as-is in the output.
Contextually, 
\citet{gaido2021moby} 
showed
that neural MT models are good at ``copying'' from the source or, in other words,
at estimating $p(Y|T)$ -- where $Y$ is the target sentence and $T$ is the textual source sentence -- when $Y$ and $T$ 
are the same string.
Hence, we hypothesize
that an ST model can close the 
performance 
gap with the ASR by conditioning the target prediction not only on the input audio,
but also on 
the generated transcript.
%
%
%
%
Formally, this means estimating $p(Y|X,T')$, where $T'$ 
denotes a representation of the generated transcript, such as the embeddings used to predict them;
and this estimation is what the triangle architecture
\cite{anastasopoulos-2018-triangle} actually does.

The triangle model is composed of a single encoder, whose output is attended by two decoders that respectively generate the transcript (ASR decoder) and the translation (ST decoder). The ST decoder also attends to the output embeddings (i.e. the internal representation before the final linear layer mapping to the output vocabulary dimension and softmax) of the ASR decoder in all its layers. In
particular, the output of the cross-attention on the encoder output and the cross-attention on the ASR decoder output are concatenated and fed to a linear layer. 
The model is optimized with a multi-loss objective function, defined as follows:

\begin{equation}
\small
\begin{aligned}
 \nonumber L(X) = - & \sum_{x \in X} \Big( \lambda_{ASR} * \sum_{t \in T_{x}} log(p_{\theta}(t_{i}|x,t_{i-1, ..., 0})) \\
 \nonumber & + \lambda_{ST} * \sum_{y \in Y_{x}} log(p_{\theta}(y_{i}|x,T,y_{i-1, ..., 0})) \Big) \label{eq:triangle_loss}
\end{aligned}
\end{equation}

where $T$ is the target transcript, $Y$ is the target translation, and $x$ is the input utterance. $\lambda_{ASR}$ and $\lambda_{ST}$ are two hyperparameters aimed at controlling the relative importance of the two tasks.
Previous works commonly set them to 0.5, giving equal importance to the two tasks \citep{anastasopoulos-2018-triangle,sperber-etal-2020-consistent}.
To the best of our knowledge, ours is the first attempt to inspect performance variations in the setting of these two parameters, calibrating them towards the specific
needs 
arising from
our application scenario.

\pgfplotstableread[row sep=\\,col sep=&]{
    ErrorType & UK & nonUKinTrain & nonUKnonTrain \\
    correct         & 52.38 & 33.33 & 36.71 \\
    misspelling     & 0.00  & 6.67  & 13.92 \\
    diff. name       & 14.29 & 26.67 & 20.25 \\
    other words            & 14.29 & 10.00 & 8.86 \\
    omission        & 19.05 & 23.33 & 20.25 \\
}\baseerrors

\pgfplotstableread[row sep=\\,col sep=&]{
    ErrorType & UK & nonUKinTrain & nonUKnonTrain \\
    correct         & 66.67 & 50.00 & 39.24 \\
    misspelling      & 0.00  & 16.67 & 12.66 \\
    diff. name       & 19.05 & 13.33 & 20.25 \\
    other words      & 9.52  & 6.67  & 17.72 \\
    omission        & 4.76  & 13.33 & 10.13 \\
}\multilingualerrors

\begin{figure*}[th]
\centering
    \footnotesize
    \subfloat[Base ASR errors.]{\label{fig:base_errors}
\begin{tikzpicture}[scale=0.95]
\begin{axis}[
            ybar,
            bar width=.2cm,
            legend style={
                at={(0.76,0.99)}, 
                anchor=north, font=\footnotesize},
            legend cell align={left},
            tickwidth         = 0pt,
            symbolic x coords={correct,misspelling,diff. name,other words,omission},
            xtick=data,
            ytick={10,20,30,40,50,60,70},
            ylabel={\%},
            ylabel near ticks,
            style={outer sep=0},
            height=4.5cm,
            width=8.4cm,
            ymin=0,
            ymax=70,
            xticklabel style={rotate=10},
            ymajorgrids=true,
        ]
        \addplot table[x=ErrorType,y=UK]{\baseerrors};
        \addplot table[x=ErrorType,y=nonUKinTrain]{\baseerrors};
        \addplot table[x=ErrorType,y=nonUKnonTrain]{\baseerrors};
        \legend{UK,non-UK in train, non-UK not in train}
    \end{axis}
\end{tikzpicture}
}
\qquad
    \subfloat[Multilingual ASR errors.]{\label{fig:multi_errors}
\begin{tikzpicture}[scale=0.95]
\begin{axis}[
            ybar,
            bar width=.2cm,
            legend style={
                at={(0.76,0.99)}, 
                anchor=north, font=\footnotesize},
            legend cell align={left},
            tickwidth         = 0pt,
            symbolic x coords={correct,misspelling,diff. name,other words,omission},
            xtick=data,
            ytick={10,20,30,40,50,60,70},
            ylabel={\%},
            ylabel near ticks,
            style={outer sep=0},
            height=4.5cm,
            width=8.4cm,
            ymin=0,
            ymax=70,
            xticklabel style={rotate=10},
            ymajorgrids=true,
        ]
        \addplot table[x=ErrorType,y=UK]{\multilingualerrors};
        \addplot table[x=ErrorType,y=nonUKinTrain]{\multilingualerrors};
        \addplot table[x=ErrorType,y=nonUKnonTrain]{\multilingualerrors};
        \legend{UK,non-UK in train, non-UK not in train}
    \end{axis}
\end{tikzpicture}
}
\caption{\label{fig:errors} Correct person names and the categories of errors of the baseline and multilingual ASR systems.}
\end{figure*}
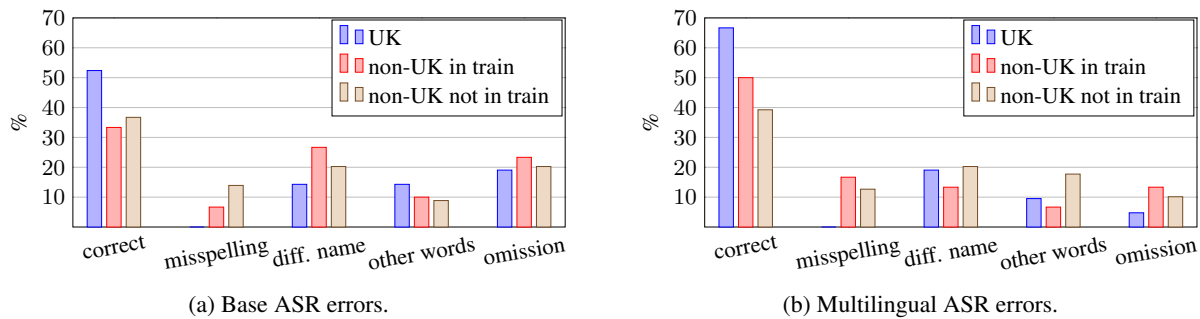

In Table \ref{tab:triangle}, we compare the multilingual models introduced in $\S\ref{sec:multiling}$ with triangle ST multilingual models trained on the same data (second row).
Although the transcripts are less accurate 
(about +1 WER), the translations have 
higher quality
(+1.4-1.6 BLEU on the three language pairs).
Person names follow a similar trend: in the transcript the accuracy is lower (-3.84), while in ST it increases (on average +2.46). Interestingly, the accuracy gap between ASR and ST is closed
by the triangle model (see the ASR-ST column),
confirming our assumption that neural models are good at copying.
However,
due to the lower ASR accuracy (42.31),
the ST accuracy (42.36) does not reach that of the base ASR model (46.15). 
The reason of this drop can be found in the different kind of information required by the ASR and ST tasks. \citet{chuang-etal-2020-worse} showed that the semantic content of the utterance is more important 
for ST,
and that
joint ASR/ST training leads the model to focus more on the semantic content of the utterance, 
yielding BLEU gains at the 
expense
of higher WER.
As person names are usually close in the semantic
space \cite{das-2017-person-embeddings}, the higher focus on semantic content may be detrimental 
to
their correct handling and hence explain the lower person name accuracy.

In light of
this observation, we experimented with changing the weights of the losses in the triangle training, 
assigning 
higher
importance to the ASR loss (third row of Table \ref{tab:triangle}). In this configuration, as expected, 
transcription
quality increases (-0.5 WER)
at the expense of
translation
quality, which decreases (-0.8 BLEU on average)
but remains higher than that of the base model.
The accuracy of person names follows the trend of 
transcription
quality:
the average accuracy on ST (46.16)
increases by
3.8 points 
over
the base triangle model (42.36),
becoming
almost identical to that of the base ASR  model (46.15). 
All in all, 
our solution achieves the same person name accuracy of an ASR base 
model without sacrificing translation quality compared to a base ST system.

\pgfplotstableread[row sep=\\,col sep=&]{
    ErrorType & UK & nonUKinTrain & nonUKnonTrain \\
    correct         & 41.18 & 34.78 & 24.59 \\
    misspelling     & 0.00  & 13.04  & 4.92 \\
    diff. name       & 17.65 & 34.78 & 31.15 \\
    other words          & 5.88 & 8.70 & 18.03 \\
    omission        & 35.29 & 8.70 & 21.31 \\
}\stbaseerrors

\pgfplotstableread[row sep=\\,col sep=&]{
    ErrorType & UK & nonUKinTrain & nonUKnonTrain \\
    correct         & 52.94 & 43.48 & 29.51 \\
    misspelling     & 0.00  & 21.74 & 9.84 \\
    diff. name       & 23.53 & 8.70 & 29.51 \\
    other words     & 11.76  & 4.35  & 14.75 \\
    omission        & 11.76  & 21.74 & 16.39 \\
}\stmultilingualerrors

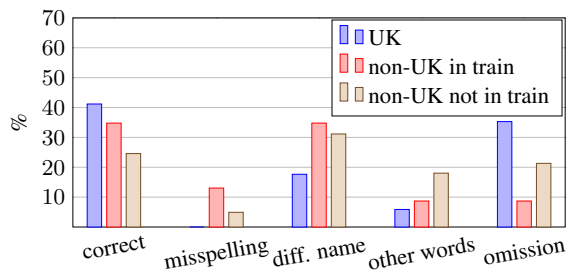
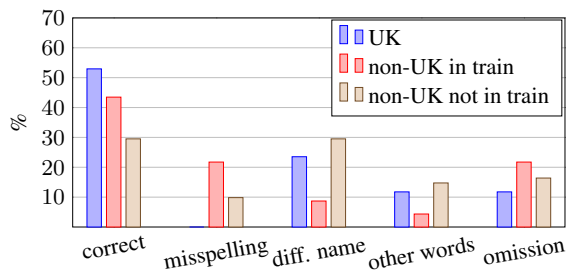
\begin{figure*}[bht]
\centering
    \footnotesize
    \subfloat[Base en-it ST errors.]{\label{fig:st_base_errors}
\begin{tikzpicture}[scale=0.95]
\begin{axis}[
            ybar,
            bar width=.2cm,
            legend style={
                at={(0.76,0.99)}, 
                anchor=north, font=\footnotesize},
            legend cell align={left},
            tickwidth         = 0pt,
            symbolic x coords={correct,misspelling,diff. name,other words,omission},
            xtick=data,
            ytick={10,20,30,40,50,60,70},
            ylabel={\%},
            ylabel near ticks,
            style={outer sep=0},
            height=4.5cm,
            width=8.4cm,
            ymin=0,
            ymax=70,
            xticklabel style={rotate=10},
            ymajorgrids=true,
        ]
        \addplot table[x=ErrorType,y=UK]{\stbaseerrors};
        \addplot table[x=ErrorType,y=nonUKinTrain]{\stbaseerrors};
        \addplot table[x=ErrorType,y=nonUKnonTrain]{\stbaseerrors};
        \legend{UK,non-UK in train, non-UK not in train}
    \end{axis}
\end{tikzpicture}
}
\qquad
    \subfloat[Multilingual ST *-it errors.]{\label{fig:st_multi_errors}
\begin{tikzpicture}[scale=0.95]
\begin{axis}[
            ybar,
            bar width=.2cm,
            legend style={
                at={(0.76,0.99)}, 
                anchor=north, font=\footnotesize},
            legend cell align={left},
            tickwidth         = 0pt,
            symbolic x coords={correct,misspelling,diff. name,other words,omission},
            xtick=data,
            ytick={10,20,30,40,50,60,70},
            ylabel={\%},
            ylabel near ticks,
            style={outer sep=0},
            height=4.5cm,
            width=8.4cm,
            ymin=0,
            ymax=70,
            xticklabel style={rotate=10},
            ymajorgrids=true,
        ]
        \addplot table[x=ErrorType,y=UK]{\stmultilingualerrors};
        \addplot table[x=ErrorType,y=nonUKinTrain]{\stmultilingualerrors};
        \addplot table[x=ErrorType,y=nonUKnonTrain]{\stmultilingualerrors};
        \legend{UK,non-UK in train, non-UK not in train}
    \end{axis}
\end{tikzpicture}
}
\caption{\label{fig:st_errors} Correct person names and the categories of errors of the baseline and multilingual ST-into-Italian systems.}
\end{figure*}

\section{Error Analysis}
\label{sec:err_analysis}

While the goal is the correct rendering of person names, not all the errors have the same weight. For interpreters, for instance, minor misspellings of a name may not be problematic, an omission
can be seen as a lack of help, but the 
generation
of a wrong name is harmful, as potentially distracting and/or confusing.
To delve into these aspects,
we first carried out a manual analysis on the ASR outputs ($\S\ref{sec:asr_err_analysis}$) and then compared the findings with the same analysis on ST outputs ($\S\ref{sec:st_err_analysis}$).

\subsection{ASR Analysis}
\label{sec:asr_err_analysis}

Two authors with at least C1 English knowledge and linguistic background annotated each error
assigning it  to a category.\footnote{The inter-annotator agreement on label assignments was calculated using the \textit{kappa coefficient} in Scott's $\pi$ formulation \citep{Scott10.1086/266577,Artstein:2008:IAC:1479202.1479206}, and resulted in 87.5\%, which means ``almost perfect'' agreement in the standard interpretation \citep{Landis77}.}
The 
categories, 
chosen by
analysing
the system outputs, are:
\textbf{misspelling} -- when a person name 
contains minor errors leading to
similar pronunciation (e.g. \textit{Kozulin} instead of \textit{Kazulin});
\textbf{replacement with a different name} -- when a person name is replaced with a completely different one in terms of spelling and/or pronunciation (e.g. \textit{Mr Muhammadi} instead of \textit{Mr Allister});
\textbf{replacement with other words} -- when a proper person name is replaced by a common noun, other parts of speech, and/or proper nouns that do not refer to people, such as geographical names (e.g. \textit{English Tibetan core} instead of \textit{Ingrid Betancourt})
\textbf{omission} -- when a person name, or part of a sentence containing it, 
is ignored by the system.

The results of the annotations are summarized in the graphs in Figure \ref{fig:errors}. Looking at the baseline system (Figure \ref{fig:base_errors}), we notice that 
omissions and replacements with a different name are the most common errors,  closely followed by replacements with other words, although for non-UK names the number of misspellings is also significant.
The multilingual system (Figure \ref{fig:multi_errors}) 
does not only show 
a higher percentage of correct names, but also a different distribution of errors, in particular for the names belonging to the languages added to the training set (non-UK in train). Indeed, the misspellings increase to the detriment of omissions and replacements with a different name and other words.
Omissions also decrease for UK names and for names 
in
languages not included in the training set 
(non-UK not in train). For
UK names, the previously-missing names fall either
into the correct names or 
into the replacements with a different name; for the non-UK not in train, instead,
they are replaced by different names or
other words.

Considering multilingual outputs, we observe that for the languages in the training set (including English), 
in 66\% of the cases the system generates a name that 
could be helpful for an interpreter (either correct or with minor misspellings).
Confusing/distracting outputs (i.e. replacements with a different person name) 
occur
in about 15\% of the cases. Future work should precisely assess whether these scores are sufficient to help interpreters in their job, or which level of accuracy is needed.

Moreover, we notice that 
the system is able to discern when a person name should be generated (either correct, misspelled, or replaced by a different name) in more than 80\% of the cases. 
This indicates their overall good capability to recognize patterns and/or appropriate contexts in which a person name should occur.

\pgfplotstableread[row sep=\\,col sep=&]{
    ErrorType & UK & nonUKinTrain & nonUKnonTrain \\
    correct         & 70.59 & 52.17 & 44.26 \\
    misspelling      & 0.00  & 8.70 & 11.48 \\
    diff. name       & 17.65 & 21.74 & 21.31 \\
    other words     & 5.88  & 0.00  & 9.84  \\
    omission        & 5.88  & 17.39 & 13.11 \\
}\sttriangleerrors

\begin{figure}[t]
\centering
    \footnotesize
\begin{tikzpicture}[scale=0.95]
\begin{axis}[
            ybar,
            bar width=.2cm,
            legend style={
                at={(0.76,0.99)}, 
                anchor=north, font=\footnotesize},
            legend cell align={left},
            tickwidth         = 0pt,
            symbolic x coords={correct,misspelling,diff. name,other words,omission},
            xtick=data,
            ytick={10,20,30,40,50,60,70},
            ylabel={\%},
            ylabel near ticks,
            style={outer sep=0},
            height=4.5cm,
            width=8.4cm,
            ymin=0,
            ymax=75,
            xticklabel style={rotate=10},
            ymajorgrids=true,
        ]
        \addplot table[x=ErrorType,y=UK]{\sttriangleerrors};
        \addplot table[x=ErrorType,y=nonUKinTrain]{\sttriangleerrors};
        \addplot table[x=ErrorType,y=nonUKnonTrain]{\sttriangleerrors};
        \legend{UK,non-UK in train, non-UK not in train}
    \end{axis}
\end{tikzpicture}
\caption{\label{fig:st_triangle_errors} Correct person names and the different categories of errors of the ST-into-Italian triangle system with $\lambda_{ASR}$=0.8, $\lambda_{ST}$=0.2 expressed in percentages.}
\end{figure}
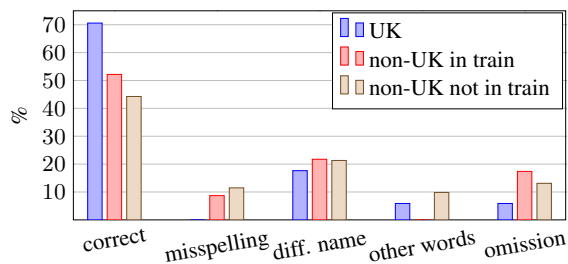

\subsection{ST Analysis}
\label{sec:st_err_analysis}

The same analysis 
was
carried out for ST systems translating into Italian (see Figure \ref{fig:st_errors}) by two native speakers, 
co-authors of this paper.
Although results are lower in general, when moving from the monolingual (Figure \ref{fig:st_base_errors}) to the multilingual (Figure \ref{fig:st_multi_errors}) system we can see similar trends to ASR, with the number of omissions and
replacements with a different name that 
decreases
in favor of a higher number of  correct names and misspellings.
Looking at the analysis of the triangle model with $\lambda_{ASR}$=0.8, $\lambda_{ST}$=0.2 presented in $\S$\ref{sec:close_gap} (Figure \ref{fig:st_triangle_errors}), we observe that misspellings, omissions, and 
replacements with other words diminish, 
while correct names increase.
Moreover, both the accuracy
(i.e. \textit{correct} in the graphs)
and the error distributions of this system are similar to those of the ASR multilingual model (Figure \ref{fig:multi_errors}).
On one side, this brings to similar conclusions, i.e. ST models can support interpreters in $\sim$66\% of the cases, and can discern when a person name is required in the translation in $\sim$80\% of the cases. On the other, it confirms
that  the gap with the ASR system is closed, as observed in $\S$\ref{sec:close_gap}.

\section{Conclusions}
Humans and machines
have different 
strengths and weaknesses.
Nonetheless, we have shown that when it comes to
person 
names
in speech, they both
struggle in 
handling
names in languages they do not know and
names that they are not used to hear.
This finding seems to insinuate
that humans cannot expect help from machines in this regard, but we demonstrated that there is hope, moving the first steps toward ST systems that can better handle person names.
Indeed, since machines are faster learners than humans, 
we can train them on more data and more languages. Moreover, we can design dedicated architectural solutions
aimed to
add an inductive bias and to improve the ability to handle specific elements.
Along this line of research,
we have shown
that a multilingual ST model, which jointly predicts the transcript and conditions the translation on it,
has relative improvements in person name accuracy by 48\% on average.
We also acknowledge
that much work is still needed in this area, with large margin of improvements available, especially to avoid 
the two most common type of errors pointed out by our analysis:
omissions and replacements with different person names.

\section*{Acknowledgement}

This work has been carried out as part of the project Smarter Interpreting (\url{https://kunveno.digital/}) financed by CDTI Neotec funds.

\bibliography{anthology,custom}
\bibliographystyle{acl_natbib}

\appendix

\section{Experimental Settings}
\label{sec:appendix}

Our ASR and ST models share the same architecture. Two 1D convolutional layers with a Gated Linear Unit non-linearity between them shrink the input sequence over the temporal dimension, having 2 as stride. Then, after adding sinusoidal positional embeddings, the sequence is encoded by 12 Transformer encoder layers, whose output is attended by 6 Transformer decoder layers.
We use 512 as Transformer embedding size, 2048 as intermediate dimension of the feed forward networks, and 8 heads.
In the case of the triangle model, we keep the same settings and the configurations are the same for the two decoders.
The number of parameters is $\sim$74M for the base system and $\sim$117M for the triangle model.

We filter out samples whose audio segment lasts more than 30s, extract 80 features from audio segments, normalize them at utterance level, and apply SpecAugment \citep{Park_2019}. The target text is segmented into
BPE  \citep{sennrich-etal-2016-neural} subwords using 8,000 merge rules \citep{di-gangi-etal-2020-target}
with SentencePience \citep{kudo-richardson-2018-sentencepiece}.

Models are optimized with Adam \cite{DBLP:journals/corr/KingmaB14} to minimize the label smoothed cross entropy \cite{szegedy2016rethinking}.
The learning rate increases up to 1e-3 for 10,000 warm-up updates, then decreases with an inverse square-root scheduler.
We train on 4 K80 GPUs with 12GB of RAM, using mini-batches containing 5,000 tokens, and accumulating the gradient for 16 mini-batches. 
We average 5 checkpoints around the best on the validation loss.
All trainings last $\sim$4 days for the multilingual systems, and $\sim$3 days for the base system.

\end{document}